\documentclass[lettersize,journal]{IEEEtran}
\usepackage{amsmath,amsfonts}
\usepackage{algorithmic}
\usepackage{algorithm}
\usepackage{array}
\usepackage{textcomp}
\usepackage{stfloats}
\usepackage{url}
\usepackage{verbatim}
\usepackage{graphicx}
\usepackage{cite}
\usepackage{booktabs} 
\usepackage{array}
\usepackage{multirow}
\usepackage{booktabs}
\usepackage{graphicx} 
\usepackage{subcaption}
\usepackage[utf8]{inputenc}          
\usepackage{newunicodechar}    
\usepackage{xcolor}
\usepackage[normalem]{ulem}
\newunicodechar{⁻}{\textsuperscript{-}}  

\newif\ifrevisionmarked
\revisionmarkedtrue
\ifrevisionmarked

\else

\fi

\begin{document}

\title{Image Bitstream Fine-grained Understanding for Privacy-Friendly AIoT}

\author{
  Zhen~Yu,  
  Wenyang~Liu,
  Kejun~Wu,~\IEEEmembership{Senior Member,~IEEE}, 
  Chengwang~Xiao,
  Renjie~Qiao,
  and Chengtao~Cai,~\IEEEmembership{Senior Member,~IEEE}
  \thanks{Zhen~Yu and Kejun~Wu are with the School of Electronic Information and Communications, Huazhong University of Science and Technology, Wuhan 430074, China.}
  \thanks{Wenyang~Liu is with the School of Electrical and Electronics Engineering, Nanyang Technological University, Singapore.}
  \thanks{Chengwang~Xiao is with School of Electronic Information, Central South University, Changsha 410083, China.}    
  \thanks{Renjie~Qiao and Chengtao~Cai are with College of Intelligent Systems Science and Engineering, Harbin Engineering University, Harbin 150001, China.}  
  \thanks{Corresponding author: Kejun~Wu (email: kjwu@hust.edu.cn).} 
  }




\maketitle

\begin{abstract}
Image Bitstream Fine-grained Understanding (IBFU) aims to directly perform fine-grained classification and semantic description generation from encoded image byte sequences. In contrast to conventional pixel-domain visual understanding, IBFU conducts semantic analysis without fully decoding images into the pixel domain. Since pixel-level visual content is not explicitly reconstructed during inference, this paradigm reduces visual exposure within the processing pipeline and suits for privacy-friendly Artificial Intelligence of Things (AIoT) applications.
In this paper, we propose Bitstream Fine-grained Generator (BFG), a novel foundation model tailored for IBFU. BFG consists of two main components: a Bitstream Semantic Encoder (BSeE) and a Fine-grained Semantic Generator (FSeG). BSeE directly models semantic representations from encoded image bitstream without explicit pixel reconstruction, while FSeG transforms the extracted bitstream semantics into detailed natural-language descriptions through autoregressive generation.
To train BFG and comprehensively evaluate IBFU in practical AIoT scenarios, where image bitstreams may suffer corruption during transmission and storage, we construct a large-scale Corrupted-bitstream Fine-grained Understanding dataset (CFU-D), containing both intact bitstreams and corrupted variants across multiple corruption types and severity levels.
Experiments show that BFG maintains stable fine-grained caption generation under bitstream corruption. For example, the performance only has slight change from 0.6339 to 0.6077 in terms of average CIDEr score on Stanford Dogs Caption dataset, while vision-language models, such as Qwen-VL-Chat, BLIP-2, GLM, Gemini, and GPT suffer severe performance decrease. This paper provides a practical paradigm for privacy-friendly fine-grained understanding in AIoT.

\end{abstract}


\section{Introduction}

\begin{figure}[!htbp]
    \centering
    \includegraphics[width=\columnwidth]{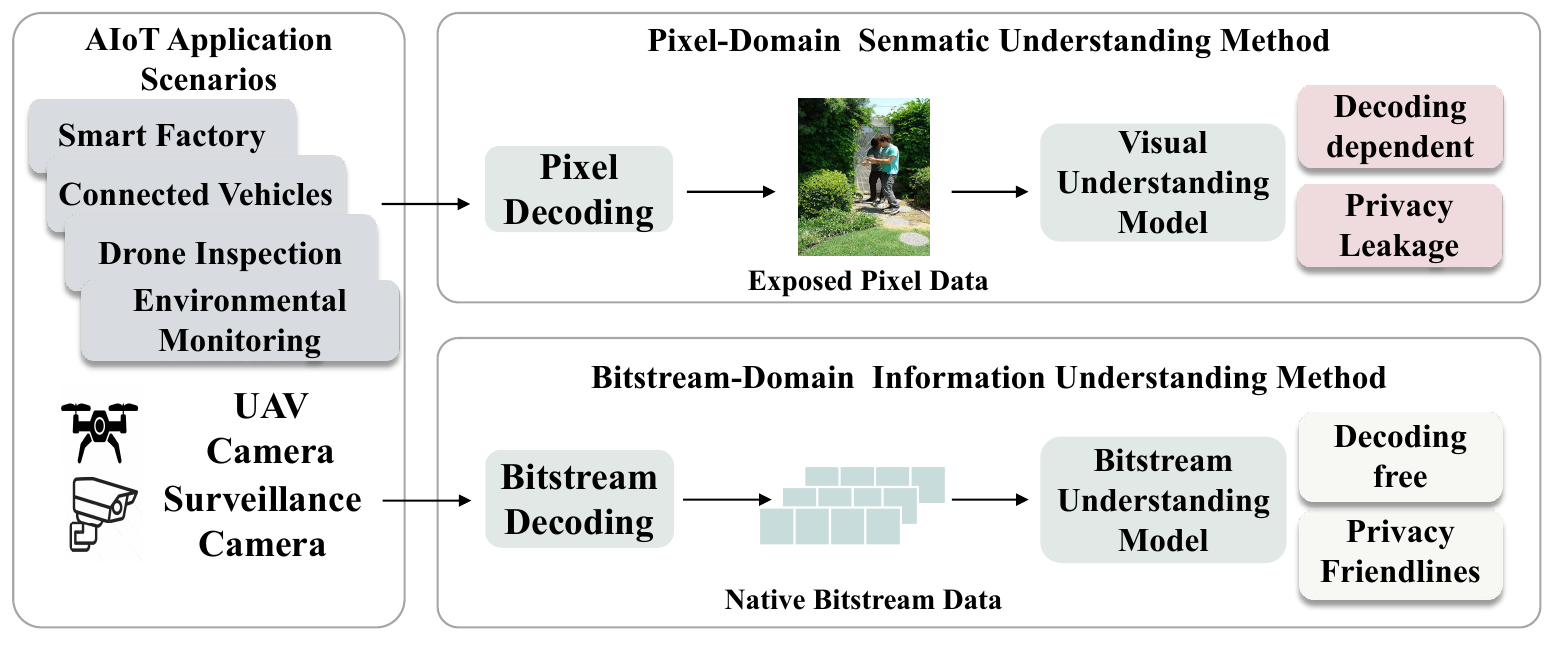}
    \caption{Bitstream-domain fine-grained understanding framework for privacy-friendly AIoT.}
    \label{fig_1}
\end{figure}

Fine-grained visual understanding task aims to distinguish subcategories that differ in subtle visual details. It supports applications in biological analysis~\cite{10506239}, human-behavior monitoring~\cite{10964313}, and medical image analysis~\cite{11052267}. For this task, convolutional neural networks and vision Transformers have achieved strong performance by learning discriminative visual representations~\cite{10870052}, while image captioning further expresses these visual cues through natural language. 
Most fine-grained vision methods are conducted on fully decoded images~\cite{wu2020block}. However, images are often stored and transmitted as compressed bitstream in real-world scenarios~\cite{wu2023end}, which may be affected by communication or storage faults~\cite{wu2024deep}. Such corruption may cause visual artifacts or prevent successful decoding~\cite{liu2025towards}. Methods that use partially decoded DCT features remain dependent on a valid JPEG structure and may become unreliable when the bitstream is structurally damaged~\cite{dong2023survey}.

ByteFormer~\cite{horton2023bytes} and bGPT~\cite{wu2024beyond} show that raw byte sequences can be modeled without conventional image decoding. More recent studies directly target corrupted image bitstreams. CBSU-ALLM treats damaged byte fragments as a specialized language for semantic classification~\cite{wu2026corrupted}, while Cibic generates captions from corrupted image bitstreams without pixel reconstruction~\cite{liang2026cibic}. These advances establish privacy-friendly feasible bitstream understanding.

Aiming at fine-grained bitstream understanding for privacy-friendly Artificial Intelligence of Things (AIoT), we propose BFG. 
In Bitstream Fine-grained Generator(BFG), Bitstream Semantic Encoder (BSeE) extracts compact local and global semantics from the byte sequence, and its Fine-grained Semantic Generator (FSeG) aligns these semantics with language tokens to produce detailed descriptions. This design supports both classification and description within a unified bitstream framework.
BFG also provides a privacy-friendly processing path for AIoT. As illustrated in Fig.~\ref{fig_1}, smart cameras and edge nodes may capture faces, identities, and private environments that become directly visible to downstream components after RGB reconstruction \cite{stanko2024artificial}. BFG performs semantic inference on encoded bytes and does not explicitly materialize an RGB image within its inference pipeline. It therefore reduces pixel-domain visual exposure while preserving the fine-grained semantics required for classification and description, making BFG suitable for privacy-friendly AIoT applications. This property concerns the processing pipeline and does not constitute a cryptographic privacy guarantee. The contributions of this paper are as follows:

\begin{itemize}
    \item We construct the Corrupted-bitstream Fine-grained Understanding dataset (CFU-D). It augments the Stanford Dogs dataset with image-level natural language descriptions and adds bitstream corruption annotations to both Stanford Dogs and Flickr30k. This dataset transforms traditional image understanding tasks into a corrupted-bitstream vision-language benchmark, enabling systematic evaluation of fine-grained classification, description quality, and robustness under bitstream corruption.
   \item We propose BFG with two coordinated modules for IBFU. BSeE extracts compact semantic representations directly from image byte sequences without pixel reconstruction, enabling bitstream-native fine-grained understanding. FSeG further combines local byte evidence with global context to generate fine-grained descriptions autoregressively. These modules allow BFG to jointly perform fine-grained classification and description on the same bitstream representation.
    \item Experimental results show that BFG maintains stable fine-grained classification and caption, while pixel-domain models suffer significant performance decrease caused by bitstream corruption. This study provides a privacy-friendly paradigm for fine-grained understanding by reducing pixel-domain visual exposure.
    
\end{itemize}

\section{Related work}
\subsection{Fine-grained Classification}
In fine-grained classification, the core difficulty lies in small inter-class differences and large intra-class variations. This requires strong local detail perception. Early research focused on part detection and local alignment, explicitly or implicitly extracting key regions of the target to capture subtle differences~\cite{al2026fine}. With the development of deep learning, weakly supervised and self-supervised methods have been widely adopted. Weakly supervised methods learn discriminative local regions using only image-level labels. Self-supervised methods such as SimCLR and MAE learn robust representations via pretext tasks, improving performance in complex scenarios~\cite{chen2020simple}. In recent years, with the development of vision-language pretraining, fine-grained research has gradually shifted toward higher-level semantic understanding and cross-modal fusion.

Models such as CLIP achieve strong cross-modal representation ability through image-text contrastive learning~\cite{radford2021learning}. Generative models such as BLIP unify visual understanding and natural language generation~\cite{li2022blip}. Meanwhile, parameter-efficient fine-tuning methods such as prompt learning and LoRA enable large models to adapt efficiently to fine-grained tasks~\cite{hu2022lora}. Recently, foundation models have attracted increasing attention. They are large-scale pre-trained general-purpose models that can adapt to various downstream tasks through fine-tuning or architectural adaptation~\cite{sun2025survey}. They have shown strong generality and effectiveness in fields such as natural language processing, computer vision, and remote sensing~\cite{awais2025foundation}. However, existing foundation models are mainly built on pixel or text modalities. Few works explore models operating directly on byte-level bitstreams. In particular, there is no foundation model that supports unified byte-sequence input while jointly handling classification and description tasks under a privacy-preserving design.

\subsection{Bitstream Understanding}
In the area of bitstream understanding, traditional visual understanding methods usually take fully decoded images as input. This leads to high computational overhead and storage requirements and makes the methods vulnerable to distortions. To reduce decoding dependency, researchers have turned to compressed-domain visual understanding. These methods directly model intermediate information generated during compression, such as DCT coefficients, quantization parameters, and JPEG block structures, to construct visual representations. This approach reduces computational complexity while preserving essential semantic information. It has been applied to tasks such as action recognition and video analysis~\cite{jamil2019optimal}. Building on this, researchers have further explored end-to-end modeling directly on raw bitstreams. Wu et al.~\cite{wu2024beyond} proposed bGPT, which represents multimodal data uniformly as byte sequences, demonstrating the feasibility of byte-level unified representation. Horton et al.~\cite{horton2023bytes} proposed ByteFormer, which takes binary byte streams directly as input and bypasses the traditional decoding pipeline through byte embedding. To address the computational overhead caused by long byte sequences, Pagnoni et al.~\cite{pagnoni2025byte} adopted a hierarchical chunking strategy to reduce complexity. Kallini et al.~\cite{kallini2025mrt5} reduced sequence length by dynamically merging similar tokens. These methods reduce reliance on full decoding and show strong unified representation capability and computational efficiency. However, most of them assume a complete and stable bitstream. When the bitstream is corrupted or structurally abnormal, the feature representations become sensitive to noise, leading to performance degradation. Improving robustness under corrupted bitstream conditions remains an open problem.

\subsection{Privacy Preservation in AIoT Systems}
Privacy has become a central concern in AIoT systems, where smart cameras and edge sensors routinely capture images containing faces, identities, and other sensitive information. The heterogeneity of these devices and their frequent communication create a broad attack surface, making unauthorized access and data leakage persistent risks~\cite{stanko2024artificial}. Various protective mechanisms have therefore been proposed. Authentication protocols and encryption techniques help secure data during transmission, while edge computing reduces the amount of sensitive information sent to the cloud by processing data locally~\cite{hou2023trends}. In smart-home environments, multi-layer security frameworks that combine device-level identity verification, link encryption, and edge-side anomaly detection have also been explored~\cite{11416471}. These measures improve the protection of data in transit and storage, but they do not fully address the exposure that may occur when visual data is analyzed.

In conventional vision pipelines, compressed images must be decoded into pixels before semantic analysis can be performed. This reconstruction creates a human-perceptible scene in system memory and makes visual details such as faces, home interiors, and personal belongings available to downstream processing components~\cite{11354649}. Data-minimization principles, including those emphasized by regulations such as the GDPR, therefore motivate approaches that limit unnecessary pixel-level exposure. Anonymization and obfuscation can hide sensitive regions, but may also remove fine-grained visual cues required for accurate recognition. Encryption protects data during storage and transmission, yet the image must generally be decrypted before pixel-based inference. Similarly, edge computing reduces transmission risks but still performs inference on decoded pixels at the local device~\cite{fu2025aiot}. These limitations have motivated research into compressed-domain and byte-level visual processing, which seeks to perform semantic analysis without explicitly reconstructing the full image. Such approaches are consistent with data-minimization and edge-processing principles, while their use for joint fine-grained classification and detailed description remains relatively underexplored.

\section{Method}
\subsection{Problem Definition}
Image Bitstream Fine-grained Understanding (IBFU) aims to recover fine-grained semantics directly from an encoded image bitstream rather than from a decoded pixel array. Let $I$ denote an image and let $\mathcal{E}(I)=B=(b_1,b_2,\ldots,b_L)$ be its encoded byte sequence, where $b_i\in\{0,\ldots,255\}$ and $L$ varies across files. Conventional visual understanding first applies an image decoder $\mathcal{D}$ to obtain a pixel tensor $X=\mathcal{D}(B)$ and then performs semantic inference on $X$. In contrast, an IBFU model takes $B$ as its only visual input and does not explicitly reconstruct $X$ during inference.

We consider two complementary IBFU outputs. Fine-grained classification estimates a category distribution $p_\theta(y\mid B)$ over visually similar subcategories. Semantic description generation predicts a token sequence $S=(s_1,\ldots,s_T)$ according to
\begin{equation}
 p_\theta(S\mid B)=\prod_{t=1}^{T}p_\theta(s_t\mid s_{<t},B),
 \label{eq:ibfu_generation}
\end{equation}
where the description expresses fine-grained attributes such as breed, appearance, pose, background, and atmosphere. IBFU must preserve these semantics not only for an intact bitstream $B$, but also for an observed bitstream $\widetilde{B}=\mathcal{C}(B;\tau,\rho)$ affected by corruption operator $\mathcal{C}$, fault type $\tau$, and severity $\rho$. The task therefore jointly concerns semantic discriminability, description fidelity, and tolerance to byte-level faults.

\begin{figure*}[!htbp]
\centering
\includegraphics[width=0.95\textwidth]{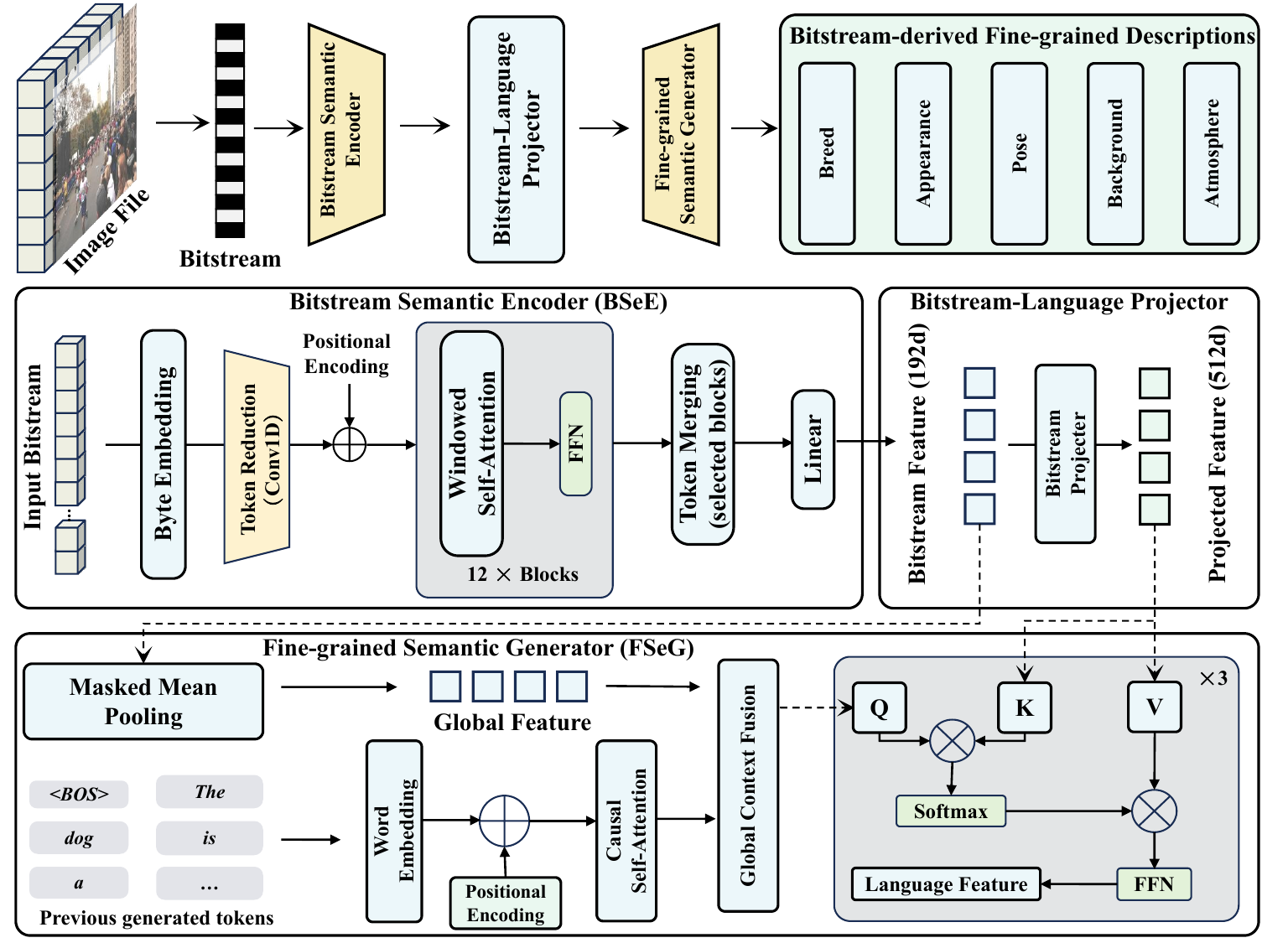}
\caption{Overall framework of BFG. BFG takes byte sequences as input and extracts bitstream features through ByteFormer, which consists of byte embedding, windowed Transformer encoding, and hierarchical sequence compression. The extracted features are then fed into a Transformer decoder with self-attention and cross-attention to autoregressively generate fine-grained semantic descriptions.}
\label{fig_2}
\end{figure*}

\subsection{BFG Architecture}
Fig.~\ref{fig_2} shows the complete forward process of BFG. Given an image file, BFG reads the file as an ordered byte sequence instead of decoding it into RGB pixels. The byte sequence first passes through BSeE to produce a sequence of 192-dimensional bitstream features. The bitstream--language projector then converts these features into 512-dimensional features compatible with the language decoder. Finally, FSeG takes the projected bitstream features and the previously generated words as its two inputs and predicts the description one token at a time. The generated description covers the breed, appearance, pose, background, and atmosphere of the image.

\subsubsection{Bitstream Semantic Encoder (BSeE)}
The input to BSeE is a JPEG bitstream $B=(b_1,b_2,\ldots,b_L)$, in which each $b_i$ is an integer byte value between 0 and 255. Because different JPEG files contain different numbers of bytes, the sequences in a batch are padded to a common length, and a binary mask marks the valid positions. The byte-embedding layer first looks up a learnable vector for every byte value, converting the discrete sequence into a continuous feature sequence. The token-reduction layer then applies one-dimensional convolution along the sequence. Each convolutional output summarizes a group of neighboring byte embeddings, so the number of tokens is reduced before the more expensive attention calculation. Positional encoding is added after token reduction to retain the order of the compressed data. This first stage is written as
\begin{equation}
 H^{(0)}=\mathcal{R}_{\mathrm{conv}}\!\left(\mathcal{E}_{b}(B)\right)+P,
 \label{eq:bsee_input}
\end{equation}
where $\mathcal{E}_{b}$ denotes byte embedding, $\mathcal{R}_{\mathrm{conv}}$ denotes Conv1D token reduction, and $P$ denotes positional encoding. The validity mask is reduced by the same stride, ensuring that padded positions are excluded from subsequent attention and pooling.

The position-aware sequence $H^{(0)}$ next enters 12 encoder blocks. In each block, windowed self-attention divides the long sequence into local windows and computes the relationships among tokens within each window. The resulting features pass through a feed-forward network, with residual connections and normalization applied around the attention and feed-forward operations. Successive blocks use shifted windows so that tokens located in neighboring windows can exchange information. At selected blocks, a token-merging operation combines adjacent tokens and shortens the sequence again. Consequently, the early blocks model short byte patterns, whereas the later blocks receive progressively larger bitstream contexts.

After the final encoder block, a linear layer maps every remaining token to a 192-dimensional representation. BSeE therefore outputs the bitstream feature sequence $Z\in\mathbb{R}^{N\times192}$ and its reduced validity mask $m\in\{0,1\}^{N}$, where $N$ is the remaining number of valid and padded tokens after reduction and merging. These features still belong to the bitstream representation space. Before sending them to the language decoder, the bitstream--language projector applies a linear transformation, CELU activation, LayerNorm, and Dropout in sequence:
\begin{equation}
 P_B=\operatorname{Dropout}\!\left(\operatorname{LN}\!\left(\operatorname{CELU}(ZW_p+b_p)\right)\right),
 \label{eq:bitstream_projection}
\end{equation}
This produces the projected feature sequence $P_B\in\mathbb{R}^{N\times512}$, which has the same feature dimension as the word representations used by FSeG. For the direct-classification pathway, BFG-Direct instead pools the valid BSeE features in $Z$ and passes the pooled vector to a 120-class linear head.

\subsubsection{Fine-grained Semantic Generator (FSeG)}
FSeG receives two input streams. The first is the projected bitstream feature sequence $P_B$ from the image file. The second is the sequence of previously available text tokens. During training, the text input is the ground-truth description shifted to the right and beginning with $\langle BOS\rangle$. During inference, it consists of $\langle BOS\rangle$ followed by the words that FSeG has already generated.

On the bitstream branch, masked mean pooling averages only the valid projected tokens and produces one 512-dimensional global feature:
\begin{equation}
 g_B=\frac{\sum_{i=1}^{N}m_iP_{B,i}}{\sum_{i=1}^{N}m_i},
 \label{eq:masked_pooling}
\end{equation}
The local sequence $P_B$ is retained because different positions may contain evidence for different visual details, while $g_B$ summarizes the complete image bitstream. Both are reused in every FSeG decoder layer.

\begin{figure*}[!htbp]
    \centering
    \includegraphics[width=\textwidth]{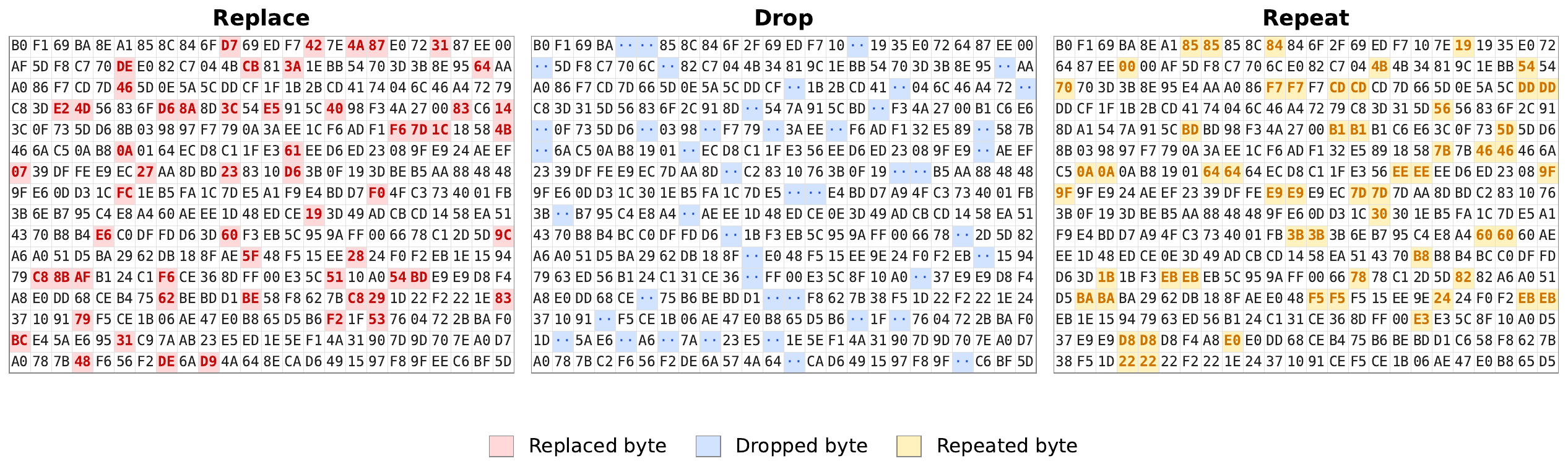}
    \caption{Illustration of corruptions on bitstream.}
    \label{fig_3}
\end{figure*}

On the text branch, each previous token is mapped to a 512-dimensional word embedding, and positional encoding is added to indicate its position in the sentence. Causal self-attention then processes this text sequence. Its causal mask allows the representation at position $t$ to attend only to tokens at positions up to $t$, preventing access to future words. The resulting language feature is combined with the global bitstream feature $g_B$ by global-context fusion. Specifically, $g_B$ is broadcast to all text positions, concatenated with each language feature, and mapped back to the decoder dimension through a fusion layer with a residual connection and normalization.

The fused language feature supplies the query $Q$ for cross-attention. On the bitstream side, the 192-dimensional BSeE output $Z$ is first converted into the 512-dimensional memory $P_B$ by the bitstream--language projector. Learned projections of this memory form the keys $K$ and values $V$ used by the decoder; thus, the dashed paths in Fig.~\ref{fig_2} indicate that both originate from the BSeE features, while the projector performs the required dimensional alignment before attention. Cross-attention is then computed as
\begin{equation}
 \operatorname{CA}(Q,K,V)=\operatorname{softmax}\!\left(\frac{QK^{\mathsf T}}{\sqrt{d}}\right)V.
 \label{eq:fseg_cross_attention}
\end{equation}
This operation assigns a weight to every valid bitstream token for each text position and returns the corresponding weighted sum of the projected values. The cross-attended representation then passes through a feed-forward network, residual connection, and normalization to produce the updated language feature. This decoder process is repeated for three stacked layers.

After the third layer, a vocabulary projection and softmax convert the language feature at the last position into probabilities for the next token. The selected token is appended to the existing text sequence and fed back into FSeG for the next decoding step. This autoregressive loop continues until $\langle EOS\rangle$ is generated or the maximum length is reached. The final output is a natural-language description rather than five independent classification scores; its sentences describe the breed, appearance, pose, background, and atmosphere.

\subsection{Construction of CFU-D}
\label{sec:cfud_construction}
We construct the Corrupted-bitstream Fine-grained Understanding dataset (CFU-D) to support both outputs of IBFU under intact and corrupted conditions. The construction pipeline contains three stages,  Dataset Composition and Description Generation, bitstream serialization, and corruption generation. Each derived sample remains linked to its source identifier and original data split so that different bitstream versions of the same image cannot cross the training and test partitions.

The source layer combines two complementary forms of supervision. Stanford Dogs provides fine-grained breed labels and is used to build the dog image classification and description subset. Its semantic target is organized as a structured natural-language description covering the breed, visual appearance, pose or activity, scene background, and overall atmosphere. To generate this target, we query the Qwen-VL-Max API with each source image and include the ground-truth breed name in the prompt. The API returns five to six sentences that begin with ``The dog is a [breed].'', where the breed name is inserted from the ground-truth folder label. We then check the returned text for the required descriptive fields, including appearance, pose or activity, scene background, and overall atmosphere, and retain the generated description as the semantic target. We manually inspected a random subset of the generated descriptions and found them to be reasonable and consistent with the source images. Breed mentions are thus anchored to the ground-truth category, while the remaining visual content is generated from the source image and verified against the required fields. Flickr30k contributes diverse scenes and human-written references, providing description supervision beyond dog images without additional API generation. Together, these sources allow CFU-D to evaluate both category-sensitive descriptions and more general semantic generation.

Bitstream serialization encodes every source image into a byte sequence and pairs it with its semantic targets. The dataset retains the intact stream as the reference input and records the valid sequence length, category label when available, textual references, source split, and encoding information. This organization separates the visual carrier (the byte sequence) from the task supervision (the class and text), enabling the same bitstream representation to support classification and generation.

Corruption generation derives corrupted variants from each intact stream using replacement, deletion, and repetition. Replacement changes selected byte values while preserving stream length, modeling value errors. Deletion removes selected bytes and shifts the following sequence, modeling information loss or incomplete transfer. Repetition duplicates selected bytes or short spans, modeling redundant insertion, retransmission, or stream-reassembly faults. Each operator is applied independently at multiple severity levels. Fig.~\ref{fig_3} illustrates these corruption patterns on the bitstream.

\section{Experiments}

\subsection{Experimental Setup}
\subsubsection{Datasets}
We construct CFU-D as the evaluation resource for IBFU. CFU-D organizes intact image bitstreams and controlled corrupted variants across replacement, deletion, and repetition faults at multiple severity levels. It contains two complementary settings. Stanford Dogs supports fine-grained recognition and description, while Flickr30k evaluates general semantic description with human-written references. We use the official Stanford Dogs split with 12,000 training images and 8,580 test images while Flickr30k evaluates general semantic description and contains 31,783 images, each with five human-written captions.Together, these two settings support a comprehensive evaluation of IBFU under both intact and corrupted bitstreams.

\begin{figure*}[htbp]
    \centering
    \includegraphics[width=\textwidth]{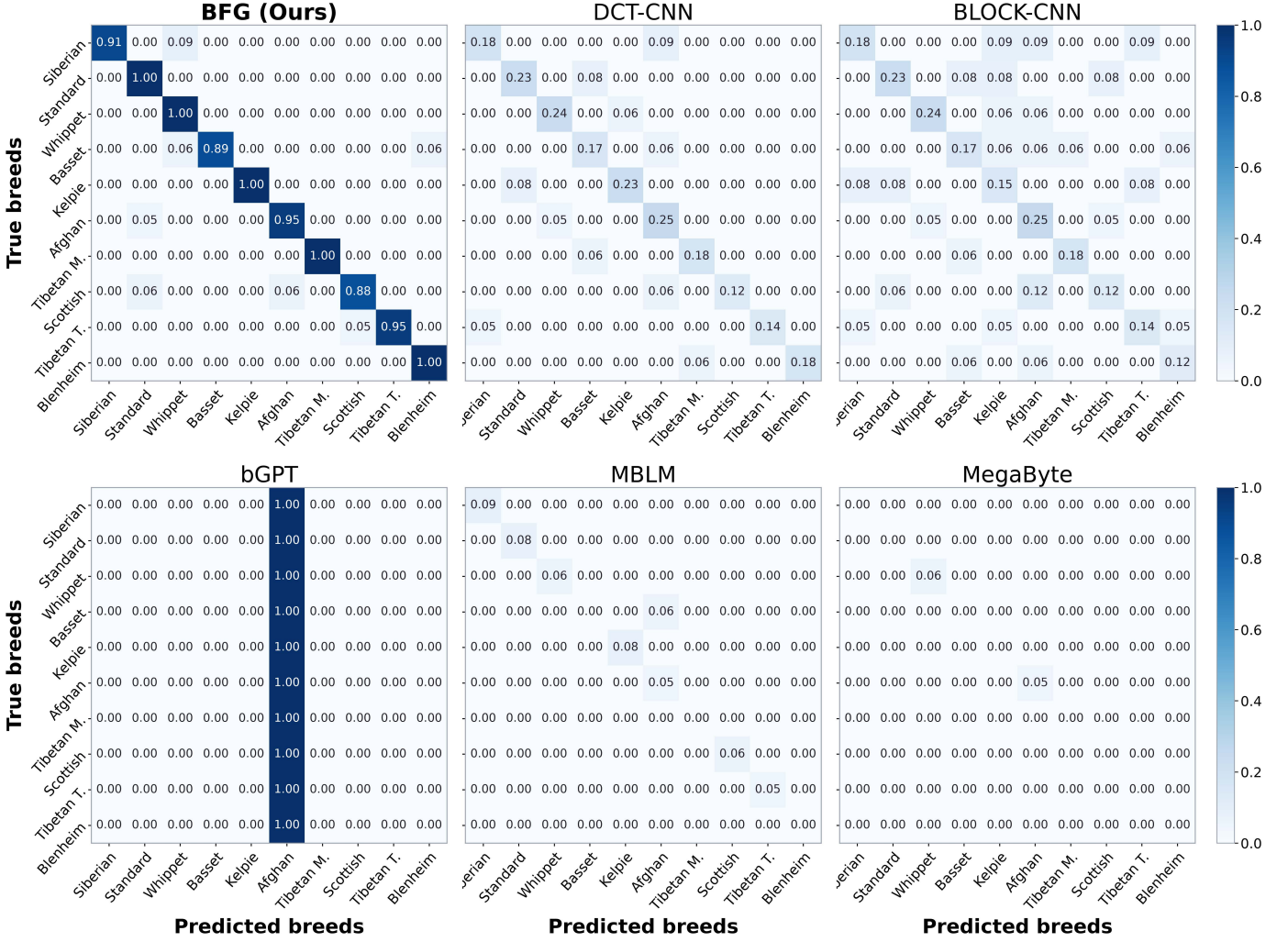}
    \caption{Confusion matrices of six bitstream-domain models on the 10 most frequently predicted Stanford Dogs breeds.}
    \label{fig:fig4}
\end{figure*}

\begin{table}[!t]
    \centering
    \caption{Comparison of classification performance of bitstream-domain models.}
    \label{tab:bitstream_cls}
    \resizebox{\columnwidth}{!}{%
    \begin{tabular}{@{}lccccc@{}}
    \toprule
    Model & Breed Acc & Precision & Recall & F1-score & Balanced Acc \\
    \midrule
   BFG-Direct & 58.51\% & 57.96\% & 58.06\% & 57.35\% & 58.06\% \\
    DCT-CNN   &  7.16\% &  5.49\% &  6.72\% &  4.35\% &  6.72\% \\
    BLOCK-CNN & 19.02\% & 17.71\% & 18.51\% & 17.59\% & 18.51\% \\
    bGPT      &  1.38\% &  0.01\% &  0.83\% &  0.02\% &  0.83\% \\
    MBLM      &  4.08\% &  3.21\% &  4.06\% &  2.93\% &  4.06\% \\
    MegaByte  &  1.39\% &  0.01\% &  0.83\% &  0.02\% &  0.83\% \\
    MambaByte &  1.20\% &  0.24\% &  1.36\% &  0.35\% &  1.36\% \\
    \bottomrule
    \end{tabular}
    }
\end{table}

\subsubsection{Implementation Details}

All experiments are implemented in PyTorch with CUDA on a single NVIDIA GPU. Images are resized to $256\times256$, center-cropped to $224\times224$, and encoded as JPEG with quality 90. Byte sequences are truncated or padded to 16,384 positions. The captioning model contains 35.257M parameters, including a frozen 15.580M-parameter ImageNet-pretrained ByteFormer and 19.677M trainable parameters; the stored weights require approximately 141 MB.

We use ByteFormer-Tiny with a 192-dimensional embedding, 12 Transformer layers, and three attention heads. The caption decoder contains three Transformer layers with a hidden dimension of 512, eight attention heads, a 2,048-dimensional feed-forward network, and dropout of 0.1. The maximum output length is 150 tokens. The model is trained for 20 epochs using Adam with a learning rate of $5\times10^{-4}$, weight decay of $10^{-4}$, batch size 16, label smoothing of 0.1, gradient clipping at 1.0, 10\% linear warm-up, and cosine decay. Caption generation uses greedy autoregressive decoding.

With batch size 1 and 16,384-byte inputs on an NVIDIA GeForce RTX 4090, the complete caption-generation pipeline requires 236.6 ms per sample on average, achieves 4.2 samples/s, and uses 223.1 MiB of peak CUDA memory. ByteFormer uses local-window attention and hierarchical sequence reduction, while decoder cost grows with the generated length.

\begin{table*}[htbp]
\centering
\caption{Performance comparison on Stanford Dogs Caption under different corruption strategies.}
\label{tab:stanford_dogs}
\small

\begin{tabular}{@{}llccccccc@{}}
\toprule
Model & Strategy & CIDEr & BLEU-4 & ROUGE-L & METEOR & BERTScore-F1 & Breed Acc & Dec.\ Rate \\
\midrule

\multirow{5}{*}{BFG(ours)} 
& No corrupt & 0.6339 & 0.0996 & 0.3651 & 0.3408 & 0.3058 & 57.63\% & - \\
& Replace    & \textbf{0.6240} & \textbf{0.0966} & \textbf{0.3623} & \textbf{0.3375} & \textbf{0.3012} & \textbf{51.24\%} & - \\
& Drop       & \textbf{0.6248} & \textbf{0.0970} & \textbf{0.3621} & \textbf{0.3376} & \textbf{0.3015} & \textbf{52.50\%} & - \\
& Repeat     & \textbf{0.5479} & \textbf{0.0748} & \textbf{0.3310} & \textbf{0.3073} & \textbf{0.2645} & \textbf{20.59\%} & - \\
& Average    & \textbf{0.6077} & \textbf{0.0920} & \textbf{0.3551} & \textbf{0.3308} & \textbf{0.2933} & \textbf{45.49\%} & - \\
\midrule

\multirow{5}{*}{Qwen-VL-Chat}
& No corrupt & 0.4389 & 0.0806 & 0.2562 & 0.0000 & 0.2118 & 38.90\% & 100.00\% \\
& Replace    & 0.0015 & 0.0146 & 0.0698 & 0.0000 & 0.0007 & 0.00\% & 34.47\% \\
& Drop       & 0.0000 & 0.0000 & 0.0000 & 0.0000 & 0.0000 & 0.00\% & 8.77\% \\
& Repeat     & 0.0089 & 0.0480 & 0.2040 & 0.0000 & 0.0043 & 0.00\% & 13.20\% \\
& Average    & 0.1123 & 0.0358 & 0.1325 & 0.0000 & 0.0542 & 9.73\% & 39.11\% \\
\midrule

\multirow{5}{*}{BLIP-2}
& No corrupt & \textbf{1.5478} & 0.1904 & 0.3565 & 0.3830 & 0.7468 & 42.61\% & 100.00\% \\
& Replace    & 0.3940 & 0.0175 & 0.0995 & 0.0892 & 0.1901 & 1.11\% & 34.47\% \\
& Drop       & 0.1089 & 0.0000 & 0.0378 & 0.0227 & 0.0525 & 0.46\% & 8.77\% \\
& Repeat     & 0.1525 & 0.0010 & 0.0483 & 0.0341 & 0.0736 & 0.47\% & 13.20\% \\
& Average    & 0.5508 & 0.0522 & 0.1355 & 0.1323 & 0.2658 & 11.16\% & 39.11\% \\
\midrule

\multirow{5}{*}{GLM-4.6V-Flash}
& No corrupt & 0.2785 & \textbf{0.2284} & \textbf{0.4162} & \textbf{0.4894} & 0.1344 & \textbf{81.71\%} & 100.00\% \\
& Replace    & 0.0248 & 0.0141 & 0.0824 & 0.0924 & 0.0120 & 1.30\% & 34.47\% \\
& Drop       & 0.0035 & 0.0001 & 0.0333 & 0.0230 & 0.0017 & 0.32\% & 8.77\% \\
& Repeat     & 0.0075 & 0.0014 & 0.0425 & 0.0358 & 0.0036 & 0.47\% & 13.20\% \\
& Average    & 0.0786 & 0.0610 & 0.1436 & 0.1602 & 0.0379 & 20.95\% & 39.11\% \\
\midrule

\multirow{5}{*}{Gemini-2.5-Flash}
& No corrupt & 0.7060 & 0.0492 & 0.2174 & 0.2157 & \textbf{0.8746} & 80.00\% & 100.00\% \\
& Replace     & 0.0000 & 0.0000 & 0.0362 & 0.0240 & 0.2836 & 0.00\% & 34.47\% \\
& Drop        & 0.0000 & 0.0000 & 0.0092 & 0.0061 & 0.0721 & 0.00\% & 8.77\% \\
& Repeat      & 0.0000 & 0.0000 & 0.0138 & 0.0092 & 0.1086 & 0.00\% & 13.20\% \\
& Average     & 0.1765 & 0.0123 & 0.0691 & 0.0637 & 0.3347 & 20.00\% & 39.11\% \\
\midrule

\multirow{5}{*}{GPT-5.1}
& No corrupt & 0.5654 & 0.0495 & 0.2083 & 0.2113 & 0.8709 & 61.11\% & 100.00\% \\
& Replace     & 0.0030 & 0.0110 & 0.0502 & 0.0433 & 0.2918 & 0.00\% & 34.47\% \\
& Drop        & 0.0008 & 0.0028 & 0.0128 & 0.0110 & 0.0742 & 0.00\% & 8.77\% \\
& Repeat      & 0.0012 & 0.0042 & 0.0192 & 0.0166 & 0.1117 & 0.00\% & 13.20\% \\
& Average     & 0.1426 & 0.0169 & 0.0726 & 0.0705 & 0.3372 & 15.28\% & 39.11\% \\

\bottomrule
\end{tabular}

\end{table*}

\subsubsection{Comparison Methods and Evaluation Metrics}
For fine-grained classification, we compare BFG-Direct with DCT-CNN, BLOCK-CNN [29], bGPT, MBLM, MegaByte, and MambaByte. BFG-Direct uses the ByteFormer encoder with a 120-class linear head and is trained only with JPEG bytes and breed labels. All methods use the official Stanford Dogs split. We report accuracy, macro precision, macro recall, macro F1-score, and balanced accuracy. We initialize BFG-Direct with ImageNet-pretrained weights, as detailed in the experimental setup.

For caption generation, BFG is compared with Qwen-VL-Chat, BLIP-2, GLM-4.6V-Flash, Gemini-2.5-Flash, and GPT-5.1. BFG processes JPEG bytes directly, whereas the pixel-domain models require JPEG decoding before inference. Qwen-VL-Chat, BLIP-2, and GLM are evaluated locally and contain 7B, 3.8B, and 9B parameters, respectively. Gemini-2.5-Flash and GPT-5.1 are proprietary API-based models with undisclosed parameter counts. Due to API cost, Gemini and GPT are evaluated on fixed subsets rather than the full test sets. We evaluate caption quality and decoding availability under the same corrupted-file conditions.

Robustness is evaluated with three reproducible byte-level fault models. Replace substitutes 2\%, 5\%, or 10\% of eligible byte values with randomly sampled values. Drop removes the same proportions of byte positions. Repeat selects 10\% or 20\% of positions and repeats each selected byte two to four times. These operations represent value corruption, information loss, and redundant insertion that may arise from transmission errors, packet loss, incomplete transfer, interrupted writes, storage damage, retransmission, or stream-reassembly faults.

Fault positions are sampled without replacement using random seed 42, while the first four JPEG signature bytes are preserved. The reported percentages indicate controlled application-level corruption severity rather than physical-layer bit-error rates. For pixel-domain models, undecodable files receive zero semantic scores, and decoding rate is reported separately. Caption quality is measured using BLEU-4, ROUGE-L, METEOR, CIDEr, and BERTScore-F1.

\subsection{Evaluations on Bitstream-Domain Fine-Grained Classification Task}

Table~\ref{tab:bitstream_cls} reports a controlled direct-classification comparison. BFG-Direct achieves 58.51\% breed accuracy on the official test split of 8,580 images. It uses a standard 120-way classification head. Compared with the other bitstream-domain models, BFG-Direct demonstrates markedly stronger fine-grained category recognition under the same evaluation setting.

To more intuitively illustrate the differences in fine-grained classification performance across models, we select the 10 most frequently predicted breeds and present the confusion matrices of all models in Fig.~\ref{fig:fig4}. BFG's confusion matrix exhibits a strong diagonal structure, with the vast majority of samples correctly classified into their corresponding breeds and only minor inter-breed confusion. The confusion matrices of BLOCK-CNN and DCT-CNN show that a large proportion of predictions fall outside the 10 target classes, with diagonal values of only 12-25\%, indicating that compression-domain methods have limited capability in fine-grained breed discrimination. bGPT's confusion matrix reveals mode collapse, with all samples predicted as afghan hound. The confusion matrices of MBLM, MegaByte, and MambaByte are extremely sparse, with nearly all predictions falling outside the 10 target breeds. These results further confirm that these models exhibit near-random discriminative ability on the 120-class fine-grained task.

\subsection{Evaluations on Robustness of Bitstream-Domain Caption Task}

\begin{table*}[htbp]
\centering
\caption{Performance comparison on Flickr30k under different corruption strategies.}
\label{tab:flickr30k}
\small

\begin{tabular}{@{}llcccccc@{}}
\toprule
Model & Strategy & CIDEr & BLEU-4 & ROUGE-L & METEOR & BERTScore-F1 & Dec.\ Rate \\
\midrule

\multirow{5}{*}{BFG(ours)} 
& No corrupt & 0.0521 & \textbf{0.1063} & \textbf{0.3005} & 0.3055 & 0.8914 & - \\
& Replace    & \textbf{0.0516} & \textbf{0.1051} & \textbf{0.2993} & \textbf{0.3021} & \textbf{0.8915} & - \\
& Drop       & \textbf{0.0636} & \textbf{0.1064} & \textbf{0.3023} & \textbf{0.3046} & \textbf{0.8926} & - \\
& Repeat     & \textbf{0.0528} & \textbf{0.1062} & \textbf{0.2977} & \textbf{0.3053} & \textbf{0.8918} & - \\
& Average    & \textbf{0.0550} & \textbf{0.1060} & \textbf{0.2999} & \textbf{0.3044} & \textbf{0.8918} & - \\
\midrule

\multirow{5}{*}{Qwen-VL-Chat}
& No corrupt & 0.0043 & 0.0196 & 0.1812 & 0.2011 & 0.8634 & 100.00\% \\
& Replace    & 0.0015 & 0.0002 & 0.0720 & 0.0415 & 0.3431 & 60.60\% \\
& Drop       & 0.0002 & 0.0000 & 0.0151 & 0.0087 & 0.0720 & 12.10\% \\
& Repeat     & 0.0003 & 0.0000 & 0.0266 & 0.0148 & 0.1267 & 21.70\% \\
& Average    & 0.0016 & 0.0049 & 0.0737 & 0.0665 & 0.3513 & 48.60\% \\
\midrule

\multirow{5}{*}{BLIP-2}
& No corrupt & \textbf{0.0547} & 0.0873 & 0.2856 & 0.2946 & \textbf{0.8939} & 100.00\% \\
& Replace    & 0.0304 & 0.0657 & 0.1756 & 0.1781 & 0.5495 & 60.60\% \\
& Drop       & 0.0055 & 0.0002 & 0.0358 & 0.0367 & 0.1121 & 12.10\% \\
& Repeat     & 0.0132 & 0.0050 & 0.0653 & 0.0660 & 0.2044 & 21.70\% \\
& Average    & 0.0259 & 0.0396 & 0.1406 & 0.1439 & 0.4400 & 48.60\% \\
\midrule

\multirow{5}{*}{GLM-4.6V-Flash}
& No corrupt & 0.0271 & 0.0099 & 0.1464 & 0.1856 & 0.8461 & 100.00\% \\
& Replace    & 0.0202 & 0.0123 & 0.0857 & 0.0978 & 0.4953 & 60.60\% \\
& Drop       & 0.0022 & 0.0002 & 0.0175 & 0.0208 & 0.1012 & 12.10\% \\
& Repeat     & 0.0061 & 0.0022 & 0.0314 & 0.0363 & 0.1815 & 21.70\% \\
& Average    & 0.0139 & 0.0062 & 0.0703 & 0.0851 & 0.4060 & 48.60\% \\
\midrule

\multirow{5}{*}{Gemini-2.5-Flash}
& No corrupt & 0.0178 & 0.1010 & 0.2509 & 0.2572 & 0.8858 & 100.00\% \\
& Replace     & 0.0081 & 0.0593 & 0.1265 & 0.1293 & 0.5305 & 60.60\% \\
& Drop        & 0.0016 & 0.0118 & 0.0252 & 0.0258 & 0.1059 & 12.10\% \\
& Repeat      & 0.0029 & 0.0212 & 0.0453 & 0.0463 & 0.1900 & 21.70\% \\
& Average     & 0.0076 & 0.0483 & 0.1120 & 0.1147 & 0.4281 & 48.60\% \\
\midrule

\multirow{5}{*}{GPT-5.1}
& No corrupt & 0.0310 & 0.1061 & 0.2649 & \textbf{0.2809} & 0.8902 & 100.00\% \\
& Replace     & 0.0084 & 0.0579 & 0.1299 & 0.0926 & 0.5335 & 60.60\% \\
& Drop        & 0.0017 & 0.0116 & 0.0259 & 0.0185 & 0.1065 & 12.10\% \\
& Repeat      & 0.0030 & 0.0207 & 0.0465 & 0.0332 & 0.1910 & 21.70\% \\
& Average     & 0.0110 & 0.0491 & 0.1168 & 0.1063 & 0.4303 & 48.60\% \\

\bottomrule
\end{tabular}

\end{table*}

We evaluate description robustness on Stanford Dogs Caption and Flickr30k in Tables~\ref{tab:stanford_dogs} and~\ref{tab:flickr30k}, respectively. Stanford Dogs Caption tests label-informed fine-grained descriptions, whereas Flickr30k tests general descriptions written by humans. On clean inputs, BLIP-2 obtains the highest Stanford Dogs CIDEr and BERTScore-F1 among the locally evaluated models, while GLM obtains the highest breed accuracy and METEOR. On Flickr30k, BFG obtains BLEU-4 of 0.1063 and ROUGE-L of 0.3005. These results show that the compact BFG model retains competitive semantic-generation capability on intact inputs.

\begin{figure*}[!htbp]
    \centering
    \includegraphics[width=0.9\textwidth]{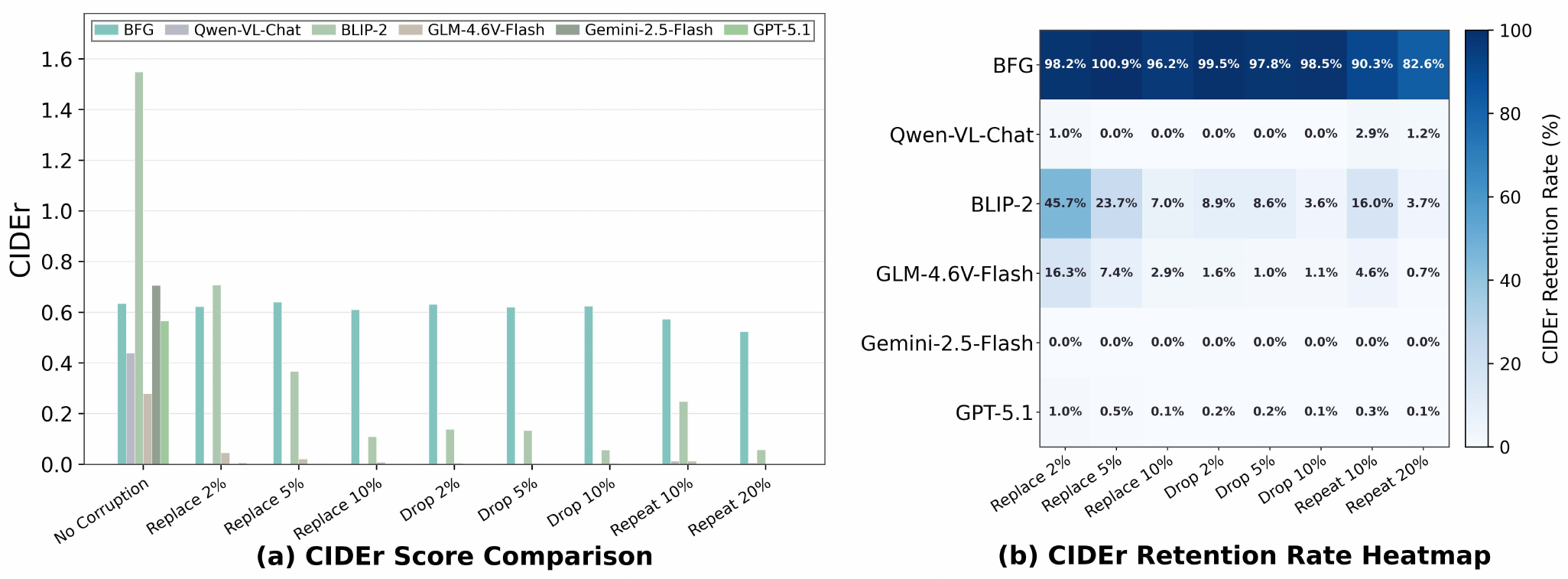}   
    \caption{CIDEr score comparison and retention rate heatmap on the Stanford Dogs Caption dataset under different corruption strategies: (a) CIDEr score comparison across models and corruption strategies; (b) CIDEr retention rate heatmap showing the percentage of baseline performance maintained.}
    \label{fig_5}
\end{figure*}

\begin{figure*}[!htbp]
    \centering
    \includegraphics[width=0.9\textwidth]{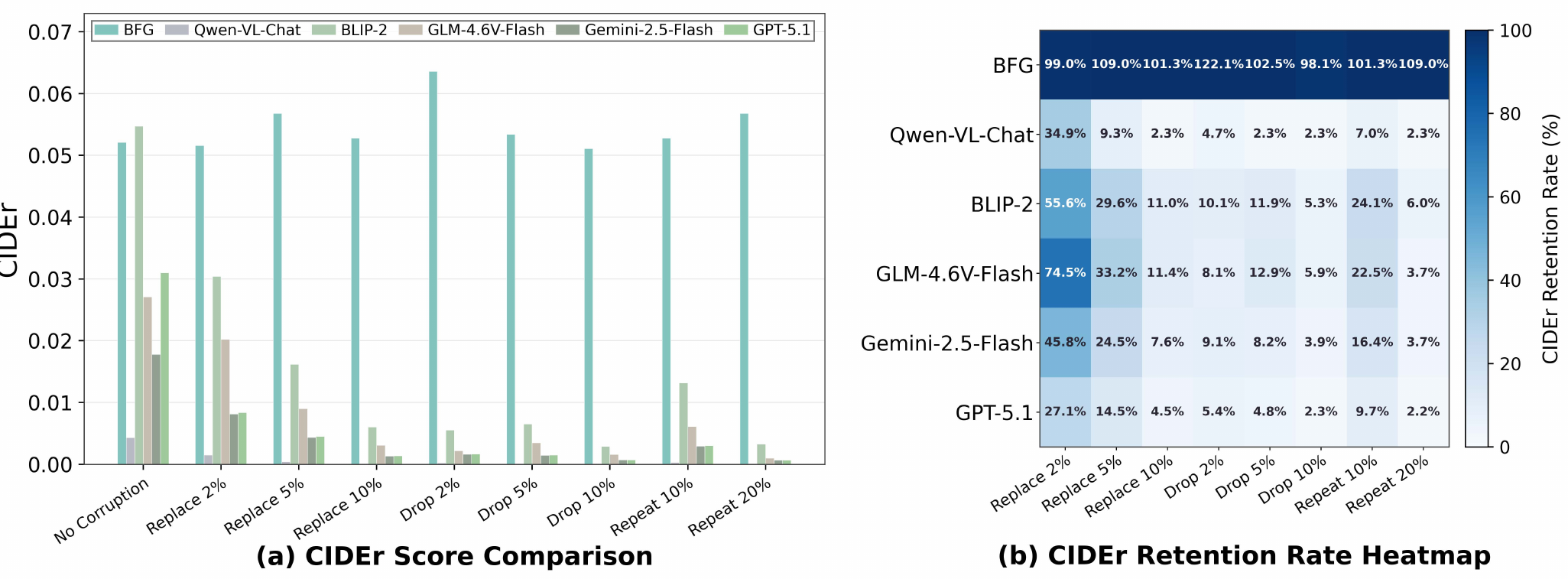}   
    \caption{CIDEr score comparison and retention rate heatmap on the Flickr30k dataset under different corruption strategies: (a) CIDEr score comparison across models and corruption strategies; (b) CIDEr retention rate heatmap showing the percentage of baseline performance maintained.}
    \label{fig_6}
\end{figure*}

\begin{figure*}[!htbp]
    \centering
    \includegraphics[width=0.94\textwidth]{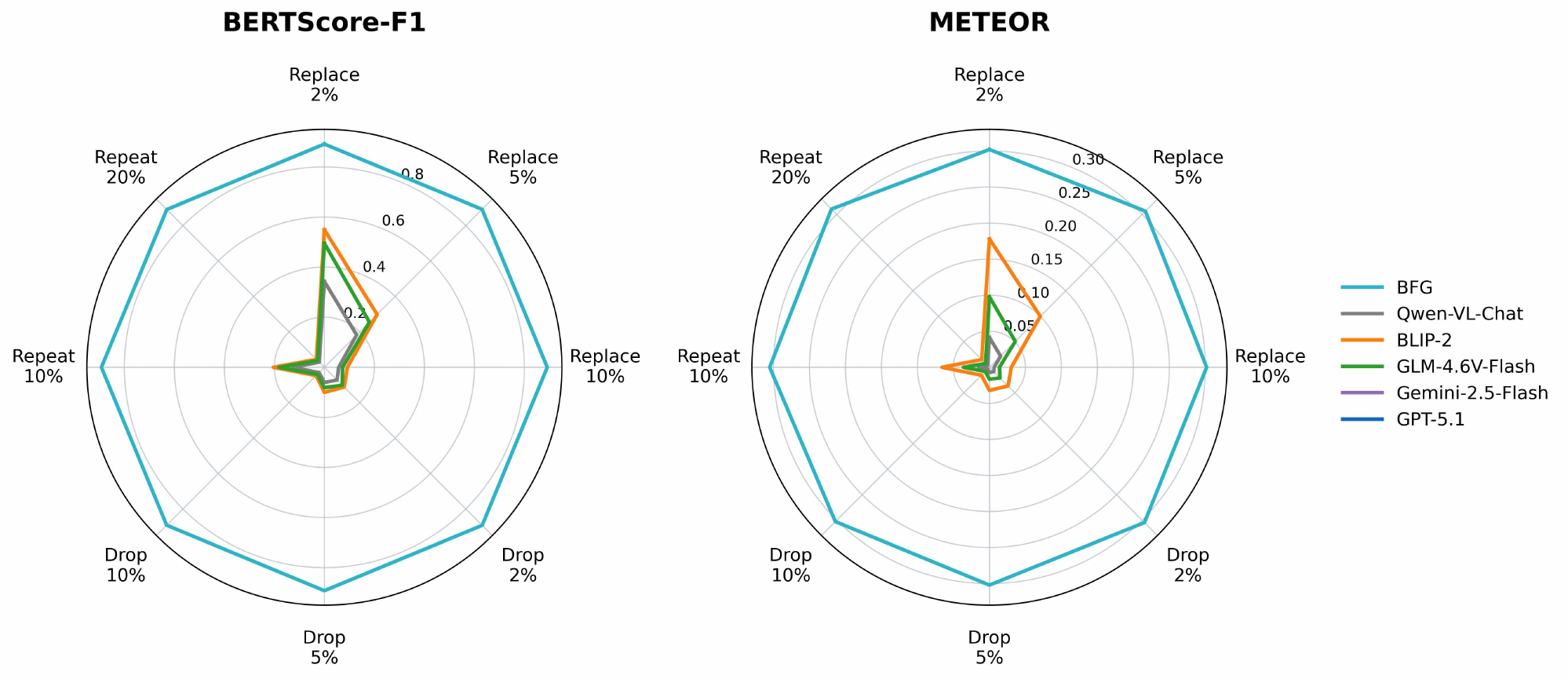}   
    \caption{Robustness comparison of BFG and vision-language models on Flickr30k. Each axis represents one corruption condition. Top row is BERTScore-F1 and bottom row is METEOR. BFG covers the largest area across all conditions.}
    \label{fig_7}
\end{figure*}

\begin{figure*}[!htbp]
    \centering
    \includegraphics[width=0.92\textwidth]{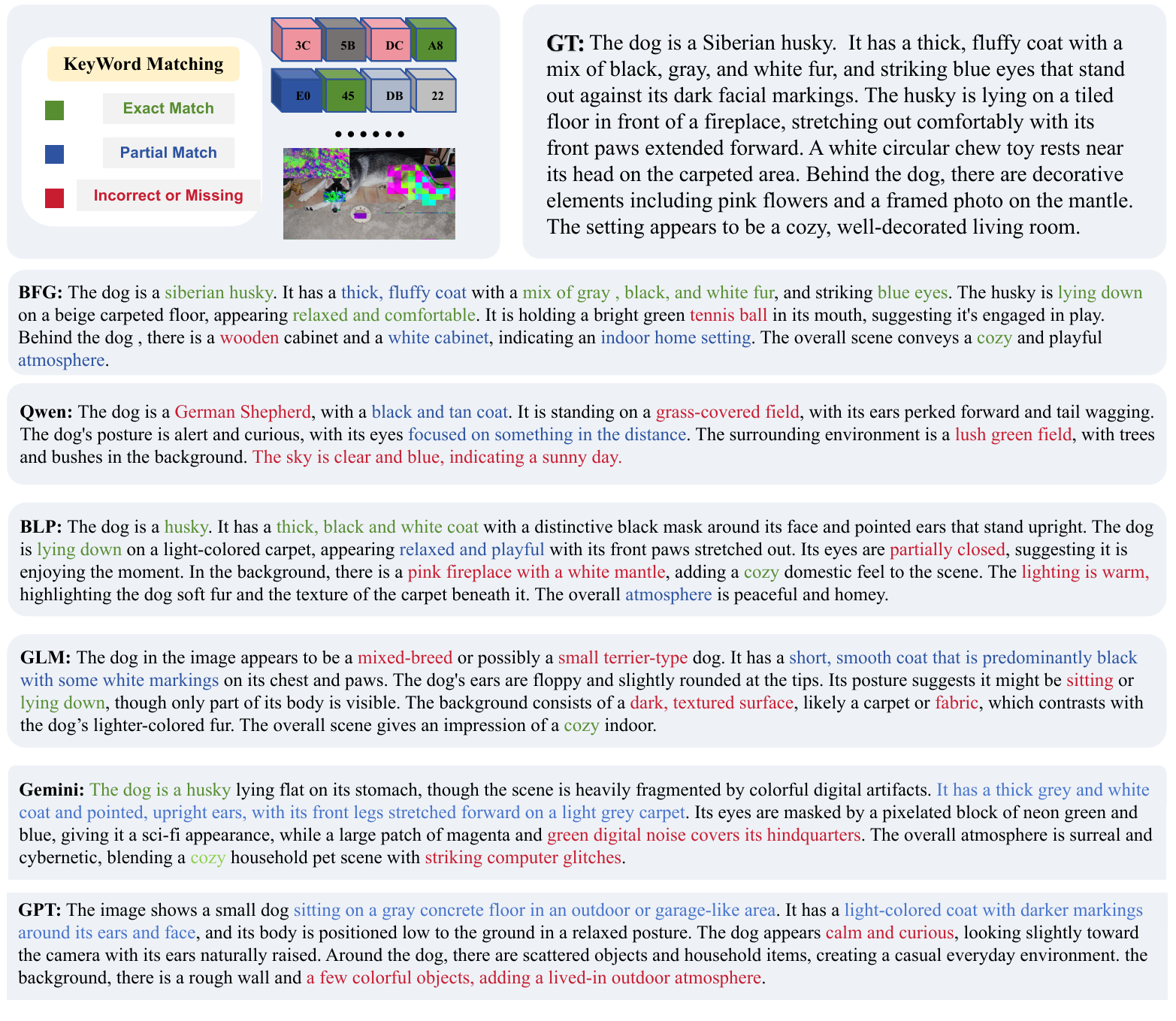}   
    \caption{Visual comparison of model-generated descriptions under 2\% byte replacement.}
    \label{fig_8}
\end{figure*}

However, once bitstream corruption is introduced, a clear performance gap emerges between the two groups of models. This is illustrated in Fig.~\ref{fig_5} and Fig.~\ref{fig_6} through CIDEr comparisons and retention rate heatmaps on the two datasets. On both datasets, under replacement and dropping strategies, BFG's CIDEr decreases by only approximately 1\% to 2\%, and the retention rates of most metrics remain above 90\%, indicating strong robustness. In contrast, vision-language models suffer severe degradation two aspects. First, corrupted bitstreams may fail to be decoded by standard decoders, leaving pixel-domain models without valid inputs. On both datasets, the decoding rates of locally deployed models (Qwen-VL-Chat, BLIP-2, GLM) decrease to approximately 60\% under 2\% byte replacement, around 10\% under 5\% byte dropping, and approach 0\% under 10\% byte dropping. API-based models (Gemini-2.5-Flash, GPT-5.1) fail entirely under any corruption. In contrast, BFG operates directly in the bitstream domain, thereby avoiding the decoding stage and maintaining complete inputs under all conditions. Second, even when decoding succeeds, the description quality of vision-language models degrades drastically. On Stanford Dogs Caption, under 5\% byte replacement, BLIP-2's CIDEr drops from 1.5478 to 0.3666 (76.3\% decrease) and breed accuracy drops from 42.61\% to 1.72\%; GLM's CIDEr drops from 0.2785 to 0.0207 (92.6\% decrease). On Flickr30k, under 5\% byte dropping, BLIP-2's CIDEr drops to 0.0055 (90\% decrease), GLM's to 0.0022 (92\% decrease), and Qwen-VL-Chat's CIDEr is near zero. Decoding distortion causes models to generate descriptions unrelated to the actual content, resulting in severe hallucinations.

Fig.~\ref{fig_7} compares the robustness of BFG and vision-language models under different corruption conditions using BERTScore-F1 and METEOR. Each radar chart has five axes corresponding to the intact bitstream condition, 2\% byte replacement, 5\% byte replacement, 2\% byte dropping, and 5\% byte dropping. On BERTScore-F1, BFG remains stable across all corruption levels. The radar area is almost unchanged. In contrast, vision-language models shrink sharply from the intact bitstream condition. METEOR shows a similar trend. These radar plots clearly reveal the performance gap between models. They also highlight the robustness advantage of bitstream-domain modeling under corruption.

\begin{table*}[!t]
\centering
\caption{Ablation on byte-level corruption augmentation for Stanford Dogs Caption. Bold values indicate the smaller relative drop for each metric and corruption condition.}
\label{tab:ablation_stanford}
\scriptsize
\resizebox{\textwidth}{!}{%
\begin{tabular}{@{}lccccc|ccccc@{}}
\toprule
& \multicolumn{5}{c}{With byte masking} & \multicolumn{5}{c}{Without byte masking} \\
\cmidrule(lr){2-6}\cmidrule(lr){7-11}
Metric & Clean & Replace & Drop & Repeat & All corrupt & Clean & Replace & Drop & Repeat & All corrupt \\
\midrule
CIDEr & 0.6339 & \textbf{0.6240} & \textbf{0.6248} & 0.5479 & \textbf{0.5989} & 0.6473 & 0.6336 & 0.6347 & \textbf{0.5637} & 0.6107 \\
BLEU-4 & 0.0996 & \textbf{0.0966} & \textbf{0.0970} & \textbf{0.0748} & \textbf{0.0895} & 0.1086 & 0.1028 & 0.1046 & 0.0720 & 0.0931 \\
ROUGE-L & 0.3651 & \textbf{0.3623} & \textbf{0.3621} & \textbf{0.3310} & \textbf{0.3518} & 0.3755 & 0.3680 & 0.3702 & 0.3262 & 0.3548 \\
METEOR & 0.3408 & \textbf{0.3375} & \textbf{0.3376} & \textbf{0.3073} & \textbf{0.3275} & 0.3506 & 0.3427 & 0.3453 & 0.3020 & 0.3300 \\
Breed Acc. & 57.63\% & \textbf{51.24\%} & 52.50\% & \textbf{20.59\%} & \textbf{41.44\%} & 62.94\% & 55.02\% & \textbf{57.44\%} & 20.18\% & 44.21\% \\
BERTScore-F1 & 0.3058 & \textbf{0.3012} & \textbf{0.3015} & \textbf{0.2645} & \textbf{0.2891} & 0.3250 & 0.3148 & 0.3176 & 0.2538 & 0.2954 \\
\bottomrule
\end{tabular}}
\end{table*}

\begin{table*}[!t]
\centering
\caption{Ablation on byte-level corruption augmentation for Flickr30k. Bold values indicate the smaller relative drop for each metric and corruption condition.}
\label{tab:ablation_flickr}
\scriptsize
\resizebox{\textwidth}{!}{%
\begin{tabular}{@{}lccccc|ccccc@{}}
\toprule
& \multicolumn{5}{c}{With byte masking} & \multicolumn{5}{c}{Without byte masking} \\
\cmidrule(lr){2-6}\cmidrule(lr){7-11}
Metric & Clean & Replace & Drop & Repeat & All corrupt & Clean & Replace & Drop & Repeat & All corrupt \\
\midrule
CIDEr & 0.0521 & 0.0537 & 0.0560 & \textbf{0.0548} & \textbf{0.0549} & 0.0498 & \textbf{0.0544} & \textbf{0.0543} & 0.0465 & 0.0517 \\
BLEU-4 & 0.1063 & \textbf{0.1060} & 0.1056 & \textbf{0.1070} & \textbf{0.1062} & 0.1046 & 0.1030 & \textbf{0.1042} & 0.0924 & 0.0999 \\
ROUGE-L & 0.3005 & \textbf{0.3004} & \textbf{0.3021} & \textbf{0.2975} & \textbf{0.3000} & 0.2956 & 0.2950 & 0.2963 & 0.2860 & 0.2925 \\
METEOR & 0.3055 & 0.3034 & 0.3031 & \textbf{0.3078} & \textbf{0.3048} & 0.3019 & \textbf{0.3003} & \textbf{0.3007} & 0.2983 & 0.2998 \\
BERTScore-F1 & 0.8914 & \textbf{0.8915} & \textbf{0.8926} & \textbf{0.8918} & \textbf{0.8920} & 0.8900 & 0.8897 & 0.8899 & 0.8868 & 0.8888 \\
\bottomrule
\end{tabular}}
\end{table*}

\begin{table}[!t]
\centering
\caption{Zero-shot evaluation on additional image formats.}
\label{tab:cross_format}
\small
\resizebox{\columnwidth}{!}{%
\begin{tabular}{@{}lcccccc@{}}
\toprule
Format & CIDEr & BLEU-4 & ROUGE-L & METEOR & BERTScore-F1 & Breed Acc. \\
\midrule
JPEG & 0.6339 & 0.0996 & 0.3651 & 0.3408 & 0.3058 & 57.63\% \\
WebP & 0.4077 & 0.0488 & 0.2913 & 0.2748 & 0.1944 & 0.44\% \\
PNG & 0.3381 & 0.0526 & 0.2934 & 0.2787 & 0.1964 & 0.44\% \\
\bottomrule
\end{tabular}}
\end{table}

\begin{figure}[!t]
    \centering
    \includegraphics[width=0.95\columnwidth]{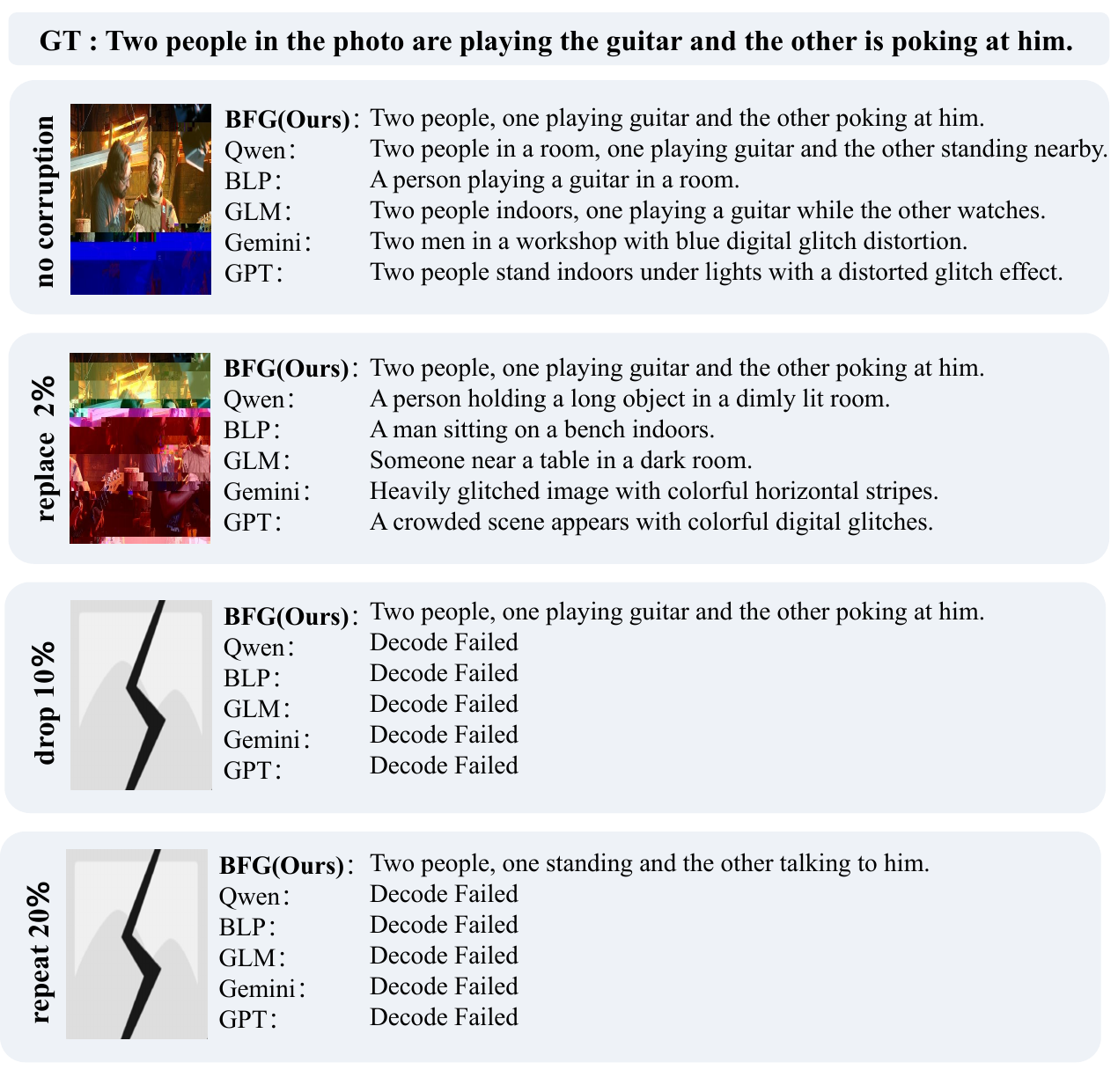}
    \caption{Visual comparison of model-generated descriptions on Flickr30k different bitstream corruption.}
    \label{fig_9}
\end{figure}

Fig.~\ref{fig_8} shows a comparison of generated captions on the Stanford Dogs Caption dataset under intact and corrupted conditions with 2\% byte replacement. Under intact conditions, all models correctly identify the breed and generate reasonable descriptions. Under corruption, the baseline vision-language models still produce outputs, but their descriptions become less accurate, often exhibiting breed misclassification and scene hallucination. In contrast, BFG remains robust and continues to produce correct breed-level descriptions with only minor attribute changes. Fig.~\ref{fig_9} presents results on the Flickr30k dataset under intact, replacement, dropping, and repetition conditions. Baseline vision-language models show noticeable degradation even under mild corruption and produce severely erroneous outputs under stronger corruption conditions due to decoding errors. BFG avoids these decoding failures and continues to generate descriptions from the corrupted bitstreams. Under the severe 20\% repetition condition, however, its output becomes less precise: the main subject remains recognizable, but the person-to-person interaction is described incorrectly. This example shows that BFG preserves input availability under severe corruption, while fine-grained semantic accuracy can still degrade when repetition disrupts the byte sequence. Overall, the results demonstrate that bitstream corruption causes both decoding failures and semantic distortion in vision-language models, whereas BFG avoids the decoding stage and remains functional under corruption.

It is worth noting that the byte repetition strategy has a larger impact on BFG compared to replacement and dropping strategies. Under repetition, BFG's breed accuracy drops more noticeably than under the other two strategies. This is because byte repetition changes the overall length and structure of the bitstream, affecting ByteFormer's sequence compression and windowed attention mechanisms. Nevertheless, BFG still maintains full input availability under all conditions, whereas the decoding rates of vision-language models under repetition do not exceed 21.7\%, and their breed accuracy is virtually 0\%.

\subsection{Ablation on byte-level corruption augmentation}
We conduct an ablation on byte-level corruption augmentation using Stanford Dogs Caption and Flickr30k, training BFG with and without this augmentation while keeping the architecture, initialization, data split, optimizer, and checkpoint-selection rule fixed. Tables~\ref{tab:ablation_stanford} and~\ref{tab:ablation_flickr} report the average results across corruption severities for each strategy, together with the overall average across corrupted inputs. Parenthesized values indicate relative changes from the corresponding clean-input score.

On Stanford Dogs Caption, byte masking reduces the average degradation for most metrics; for example, the overall BERTScore-F1 drop decreases from 9.10\% to 5.47\%. On Flickr30k, its clearest benefit appears in the average results under repetition. Overall, the ablation shows that byte masking improves caption stability under corrupted inputs.

\subsection{Zero-shot Generalization to Other Image Formats}
We use the cross-format experiment in Table~\ref{tab:cross_format} to assess whether BFG's bitstream understanding extends beyond JPEG. The results show substantially lower caption scores on WebP and PNG, with breed accuracy of 0.44\% on both formats. Thus, BFG's current generalization is limited to JPEG bitstreams.

\section{Conclusion}
This paper introduces IBFU and presents BFG for fine-grained classification and semantic description directly from encoded image bitstreams. BSeE extracts compact semantic representations from JPEG bytes, while FSeG combines local evidence with global context to generate detailed descriptions. We also construct CFU-D, which contains intact bitstreams and corrupted variants across multiple corruption types and severity levels for evaluating fine-grained classification, description quality, and robustness. Experiments show that BFG maintains stable fine-grained caption generation under bitstream corruption, whereas pixel-domain vision-language models suffer severe performance degradation or decoding failure. In addition, BFG reduces pixel-domain visual exposure within the inference pipeline, offering a practical pathway for privacy-friendly fine-grained understanding in AIoT.

\bibliographystyle{IEEEtran}
\bibliography{ref}

@article{wu2020block,
  title={Block-wise focal stack image representation for end-to-end applications},
  author={Wu, Kejun and Yang, You and Yu, Mei and Liu, Qiong},
  journal={Optics Express},
  volume={28},
  number={26},
  pages={40024--40043},
  year={2020},
  publisher={Optical Society of America}
}

@article{wu2023end,
  title={End-to-end varifocal multiview images coding framework from data acquisition end to vision application end},
  author={Wu, Kejun and Liu, Qiong and Wang, Yi and Yang, You},
  journal={Optics Express},
  volume={31},
  number={7},
  pages={11659--11679},
  year={2023},
  publisher={Optica Publishing Group}
}

@article{wu2024deep,
  title={Deep video compression based on long-range temporal context learning},
  author={Wu, Kejun and Li, Zhenxing and Yang, You and Liu, Qiong},
  journal={Computer Vision and Image Understanding},
  volume={248},
  pages={104127},
  year={2024},
  publisher={Elsevier}
}

@inproceedings{liu2025towards,
  title={Towards blind bitstream-corrupted video recovery: A visual foundation model-driven framework},
  author={Liu, Tianyi and Wu, Kejun and Cai, Chen and Wang, Yi and Yap, Kim-Hui and Chau, Lap-Pui},
  booktitle={Proceedings of the 33rd ACM International Conference on Multimedia},
  pages={7949--7958},
  year={2025}
}

@article{wu2026corrupted,
  title={Corrupted bitstream semantic understanding by adaptive-modal large language models},
  author={Wu, Kejun and Li, Fangcheng and Liu, Wenyang and Liu, Qiong and Yang, You},
  journal={Pattern Recognition},
  volume={180},
  pages={114151},
  year={2026},
  publisher={Elsevier}
}

@article{liang2026cibic,
  title={Cibic: Pixel-free foundation model for robust corrupted image bitstream captioning},
  author={Liang, Junnan and Wu, Kejun and Hu, Xuanwei and Liu, Tianyi},
  journal={Pattern Recognition},
  pages={114238},
  year={2026},
  publisher={Elsevier}
}

@article{dong2023survey,
  title={A survey on compression domain image and video data processing and analysis techniques},
  author={Dong, Yuhang and Pan, W David},
  journal={Information},
  volume={14},
  number={3},
  pages={184},
  year={2023},
  publisher={MDPI}
}

@article{horton2023bytes,
  title={Bytes are all you need: Transformers operating directly on file bytes},
  author={Horton, Maxwell and Mehta, Sachin and Farhadi, Ali and Rastegari, Mohammad},
  journal={arXiv preprint arXiv:2306.00238},
  year={2023}
}

@article{wu2024beyond,
  title={Beyond language models: Byte models are digital world simulators},
  author={Wu, Shangda and Tan, Xu and Wang, Zili and Wang, Rui and Li, Xiaobing and Sun, Maosong},
  journal={arXiv preprint arXiv:2402.19155},
  year={2024}
}

@inproceedings{stanko2024artificial,
  title={Artificial Intelligence of Things (AIoT): Integration Challenges, and Security Issues.},
  author={Stanko, Andrii and Duda, Oleksii and Mykytyshyn, Andrii and Totosko, Oleg and Koroliuk, Rostyslav},
  booktitle={BAIT},
  pages={92--105},
  year={2024}
}

@article{hou2023trends,
  title={Trends and challenges in AIoT/IIoT/IoT implementation},
  author={Hou, Kun Mean and Diao, Xunxing and Shi, Hongling and Ding, Hao and Zhou, Haiying and De Vaulx, Christophe},
  journal={Sensors},
  volume={23},
  number={11},
  pages={5074},
  year={2023},
  publisher={MDPI}
}

@article{al2026fine,
  title={Fine-grained Visual Classification: A Survey of Methods, Architectures, and Emerging Trends},
  author={Al Amin, Md and Jung, Ho Yub},
  journal={IEEE Access},
  year={2026},
  publisher={IEEE}
}

@inproceedings{chen2020simple,
  title={A simple framework for contrastive learning of visual representations},
  author={Chen, Ting and Kornblith, Simon and Norouzi, Mohammad and Hinton, Geoffrey},
  booktitle={International conference on machine learning},
  pages={1597--1607},
  year={2020},
  organization={PmLR}
}

@inproceedings{radford2021learning,
  title={Learning transferable visual models from natural language supervision},
  author={Radford, Alec and Kim, Jong Wook and Hallacy, Chris and Ramesh, Aditya and Goh, Gabriel and Agarwal, Sandhini and Sastry, Girish and Askell, Amanda and Mishkin, Pamela and Clark, Jack and others},
  booktitle={International conference on machine learning},
  pages={8748--8763},
  year={2021},
  organization={PmLR}
}

@inproceedings{li2022blip,
  title={Blip: Bootstrapping language-image pre-training for unified vision-language understanding and generation},
  author={Li, Junnan and Li, Dongxu and Xiong, Caiming and Hoi, Steven},
  booktitle={International conference on machine learning},
  pages={12888--12900},
  year={2022},
  organization={PMLR}
}

@article{hu2022lora,
  title={Lora: Low-rank adaptation of large language models.},
  author={Hu, Edward J and Shen, Yelong and Wallis, Phillip and Allen-Zhu, Zeyuan and Li, Yuanzhi and Wang, Shean and Wang, Liang and Chen, Weizhu and others},
  journal={Iclr},
  volume={1},
  number={2},
  pages={3},
  year={2022}
}

@article{sun2025survey,
  title={A survey of reasoning with foundation models: Concepts, methodologies, and outlook},
  author={Sun, Jiankai and Zheng, Chuanyang and Xie, Enze and Liu, Zhengying and Chu, Ruihang and Qiu, Jianing and Xu, Jiaqi and Ding, Mingyu and Li, Hongyang and Geng, Mengzhe and others},
  journal={ACM Computing Surveys},
  volume={57},
  number={11},
  pages={1--43},
  year={2025},
  publisher={ACM New York, NY}
}

@article{awais2025foundation,
  title={Foundation models defining a new era in vision: a survey and outlook},
  author={Awais, Muhammad and Naseer, Muzammal and Khan, Salman and Anwer, Rao Muhammad and Cholakkal, Hisham and Shah, Mubarak and Yang, Ming-Hsuan and Khan, Fahad Shahbaz},
  journal={IEEE Transactions on Pattern Analysis and Machine Intelligence},
  volume={47},
  number={4},
  pages={2245--2264},
  year={2025},
  publisher={IEEE}
}

@article{jamil2019optimal,
  title={An optimal codebook for content-based image retrieval in JPEG compressed domain},
  author={Jamil, Afshan and Majid, Muhammad and Anwar, Syed Muhammad},
  journal={Arabian Journal for Science and Engineering},
  volume={44},
  number={11},
  pages={9755--9767},
  year={2019},
  publisher={Springer}
}

@inproceedings{pagnoni2025byte,
  title={Byte latent transformer: Patches scale better than tokens},
  author={Pagnoni, Artidoro and Pasunuru, Ramakanth and Rodriguez, Pedro and Nguyen, John and Muller, Benjamin and Li, Margaret and Zhou, Chunting and Yu, Lili and Weston, Jason E and Zettlemoyer, Luke and others},
  booktitle={Proceedings of the 63rd Annual Meeting of the Association for Computational Linguistics (Volume 1: Long Papers)},
  pages={9238--9258},
  year={2025}
}

@inproceedings{kallini2025mrt5,
  title={Mrt5: Dynamic token merging for efficient byte-level language models},
  author={Kallini, Julie and Murty, Shikhar and Manning, Christopher and Potts, Christopher and Csord{\'a}s, R{\'o}bert},
  booktitle={International Conference on Learning Representations},
  volume={2025},
  pages={56646--56669},
  year={2025}
}

@INPROCEEDINGS{11416471,
  author={Sharma, Dilli Prasad and Sun, Xiaowei and Xue, Liang and Lin, Xiaodong and Xiong, Pulei},
  booktitle={2025 IEEE Annual Congress on Artificial Intelligence of Things (AIoT)}, 
  title={Privacy-Preserving Explainable AIoT Application via SHAP Entropy Regularization}, 
  year={2025},
  volume={},
  number={},
  pages={253-261},
  doi={10.1109/AIoT66900.2025.00046}}

@INPROCEEDINGS{11354649,
  author={Zhang, Can},
  booktitle={2025 IEEE 24th International Conference on Trust, Security and Privacy in Computing and Communications (TrustCom)}, 
  title={DPDA: Decentralized Identity-Based Privacy-Preserving Data Aggregation for AIoT}, 
  year={2025},
  volume={},
  number={},
  pages={3243-3249},
  doi={10.1109/Trustcom66490.2025.00390}}

@article{fu2025aiot,
  title={AIoT: Artificial Intelligence and the Internet of Things for monitoring and prognosis of systems and structures},
  author={Fu, Hailing and Rao, Jing and Deng, Fang and Wang, Yihan and Zhao, Bowen and Liu, Zhuowen and Guan, Hong and Malinowski, Pawe{\l} H and Xu, Lijun},
  journal={IEEE Transactions on Instrumentation and Measurement},
  year={2025},
  publisher={IEEE}
}

@ARTICLE{10506239,
  author={Lu, Qitao and Xia, Guoyuan and Cai, Mingjing and Li, Xin and Cao, Junyi and Liao, Wei-Hsin},
  journal={IEEE Internet of Things Journal}, 
  title={An Ultralow Frequency Energy Harvester With an Asymmetric-Stiffness Pendulum Inspired by Biological Grooming Behavior}, 
  year={2024},
  volume={11},
  number={13},
  pages={24222-24233}
  }

@ARTICLE{10964313,
  author={Singh, Akshay and Thakur, Rahul},
  journal={IEEE Internet of Things Journal}, 
  title={A Cognitive-Intelligence-Based Personalized Federated Approach for Monitoring Driver Behavior}, 
  year={2025},
  volume={12},
  number={14},
  pages={26116-26127}
  }

@ARTICLE{11052267,
  author={Liu, Jinlei and Yu, Fuyao and Li, Rui and Lyu, Xiaohong and Zheng, Shilei},
  journal={IEEE Internet of Things Journal}, 
  title={Enhancement and Segmentation of High Definition CT Images in Everything 6G Medical IoT Environment}, 
  year={2026},
  volume={13},
  number={5},
  pages={7862-7873}
  }

@ARTICLE{10870052,
  author={Ma, Shuai and Sun, Zhiye and Shen, Bin and Wu, Youlong and Li, Hang and Shi, Guangming and Li, Shiyin and Al-Dhahir, Naofal},
  journal={IEEE Internet of Things Journal}, 
  title={Semantic Feature Division Multiple Access for Digital Semantic Broadcast Channels}, 
  year={2025},
  volume={12},
  number={11},
  pages={17123-17136}
  }

\end{document}